\documentclass[runningheads]{llncs}
\usepackage{graphicx}
\usepackage{amsmath,amssymb} %
\usepackage{color}

\begin{document}
\title{Brand $>$ Logo: Visual Analysis of Fashion Brands} 

\titlerunning{Brand $>$ Logo: Visual Analysis of Fashion Brands}

\author{M. Hadi Kiapour \and
Robinson Piramuthu}

\authorrunning{M. Hadi Kiapour and Robinson Piramuthu}

\institute{eBay, San Francisco CA 94105, USA\\
\email{\{mkiapour,rpiramuthu\}@ebay.com}}
\maketitle              %
\begin{abstract}
While lots of people may think branding begins and ends with a logo, fashion brands communicate their uniqueness through a wide range of visual cues such as color, patterns and shapes. In this work, we analyze learned visual representations by deep networks that are trained to recognize fashion brands. In particular, the activation strength and extent of neurons are studied to provide interesting insights about visual brand expressions. The proposed method identifies where a brand stands in the spectrum of branding strategy, i.e., from trademark-emblazoned goods with bold logos to implicit no logo marketing. By quantifying attention maps, we are able to interpret the visual characteristics of a brand present in a single image and model the general design direction of a brand as a whole. We further investigate versatility of neurons and discover ``specialists" that are highly brand-specific and ``generalists" that detect diverse visual features. A human experiment based on three main 
visual scenarios of fashion brands is conducted to verify the alignment of our quantitative measures with the human perception of brands. This paper demonstrate how deep networks go beyond logos in order to recognize clothing brands in an image.

\end{abstract}
\section{Introduction}
On your walk home, a runner whisks past you. Her feet flying over the concrete and leaves, they make a blur of a small but unmistakable check mark. This remarkably simple logo, dubbed the swoosh, perfectly embodies motion and speed, attributes of the winged Goddess of victory in Greek mythology, Nike.

Fashion is all about identity. From luxury splurges to mass retail sneakers, logos have been considered a key status symbol. Over time however, as buying habits change, the status symbols evolve. Since the rise of the No Logo movement~\cite{nologo}, some brands have embraced minimalism. Louis Vuitton made news in 2013 when it pulled back on the use of its iconic LVs on purses.
Good branding is more than a logo. It is storytelling; a visual story woven into every piece. Here's a test: if you cover up the logo on a product, can you still tell the brand?

\begin{figure}[t]
\centering
\includegraphics[width=1.0\textwidth]{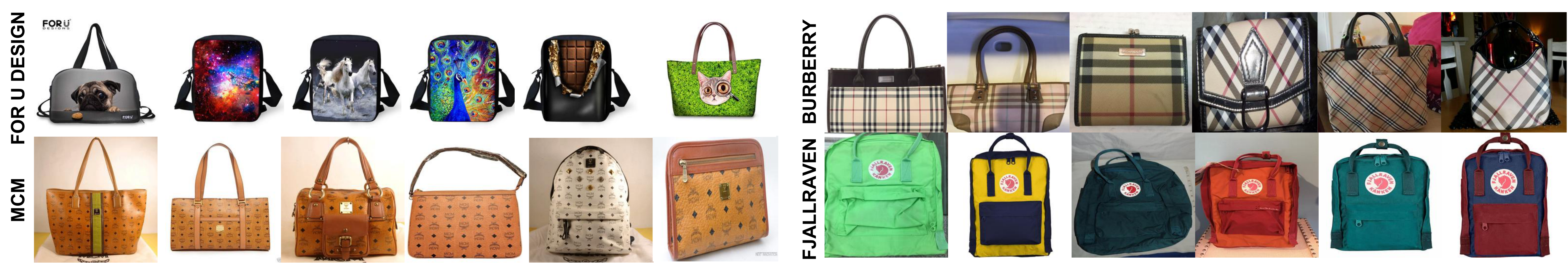}
\caption{Visual brands spectrum. Brands use a wide range of visual expressions. Experts can identify brands even in the absence of logo. While some use colorful graphics, others adopt unique patterns or choose to rely on logo.}
\label{fig:introduction_different_brand_strategy}
\end{figure}

Uniqueness is a vital factor for a successful clothing business. Shoppers not only want to be fashionable, but also want to express themselves. In 1992 Christian Louboutin decided to create a signature style that hints at sensuality and power simultaneously, and they painted the soles of their shoes red! There are a million ways that designers make memorable brand expressions. Sometimes they bring life to a logo, other times they use patterns to make a brand recognizable. Some make eccentric products in shape and geometry and others make name for themselves by unique color combinations, folds and cuts. Figure~\ref{fig:introduction_different_brand_strategy} shows examples drawn from the wide spectrum of visual expressions fashion brands adopt. While some use colorful graphics or repeated logo prints, others design unique patterns or mainly invest on logos.

With the recent success of computer vision and the rise of online commerce, there is a huge excitement to turn computers into visual experts. The ever-changing landscape of fashion industry has provided a unique opportunity to leverage computational algorithms on large data to achieve the knowledge and expertise unattainable for any individual fashion expert. Previous researchers have worked on clothing parsing~\cite{yamaguchi:paperdoll,gong:lookinto}, outfit compatibility and recommendation~\cite{veit:learningvisual,yamaguchi:mix}, style and trend recognition~\cite{kiapour:hipsterwars,alhalah:fashionforward}, attribute recognition~\cite{chen:deepdomain,chen:describing,liu:fashionlandmark} and retrieval~\cite{kiapour:wtb,liu:deepfashion}. In order to interpret deep visual representations, studies have discovered neurons that can predict semantic attributes shared among categories~\cite{ozeki:understandingconvolutional,escorcia:ontherelationship} and grand-mother-cell like features~\cite{agrawal:analyzingtheperformance} and probed the neuron activations to discover concepts~\cite{vittayakorn:automaticattribute,bau:networkdissection,zhou:objectdetectors}. Another body of research rely on attention paradigm to find parts of the image that are most responsible for the classification~\cite{ruth:interpretable,selvaraju:gradcam,zeiler:visualizing,springenberg:striving}. Our work builds upon the top-down attention mechanism of Zhang et al.~\cite{zhang:topdown} to uncover what computer vision models learn in order to distinguish fine-grained fashion brands across a wide variety of products. Specifically, we aim to answer the following questions:
\begin{itemize}
\item How can we quantify visual brand representations?
\item How do deep networks distinguish between very similar products?
\item What are the key visual expressions that brands adopt?
\item Which visual representations are shared or unique across brands?
\item How well does the learned representations align with human perception?
\end{itemize}

\section{Methodology}
\noindent
\textbf{Data}. We collect a new large dataset of $3,828,735$ clothing product images from $1219$ brands taken from a global online marketplace reported in Table~\ref{tab:data_stats}. The dataset contains diverse images from stock quality photos taken professionally with white background to photos of used products photographed by amateurs in challenging viewpoints and lighting. We grouped the products to fall into five broad categories:\emph{Bags, Footwear, Bottom wear, Outerwear} and \emph{Tops}.

\noindent
\textbf{Classification Network}.
In deep learning, fine-tuning a convolutional network, pretrained on large data, is considered as a simple transfer learning to provide good initialization~\cite{transferlearning}. We fine-tune the ResNet-50 model on ImageNet~\cite{he:deepresidual,krizhevsky:imagenet} for classification among the $1219$ brands in our dataset and achieve $47.1\%$ top-1 accuracy. Next we use an attention mechanism to generate brand-specific attention maps on the convolution layers. In our experiments, we study \emph{res5b} maps due to its manageable size and proximity to the final classification layer. Our method can be applied to any convolution layer in deep networks.

\begin{table}
\centering
\caption{Fashion brands dataset collected from online e-commerce sites.}
\begin{tabular}{llccc}
\hline
\textbf{Category} & \textbf{Subcategories} & \textbf{\#Brands} & \textbf{\#Train} & \textbf{\#Test}\\
\hline
Bags       & Handbags, Purses       & $132$      &  $206,232$  & $18,427$  \\ \hline
                            
Footwear       & Shoes, Heels, Boots       & $235$      & $368,846$  & $37,825$  \\ \hline
                            
Bottom wear       & Pants, Jeans, Skirts       & $218$      &  $431,568$  & $44,408$  \\ \hline
                            
Outer wear       & Jackets, Coats       & $238$      &  $442,950$  & $42,962$  \\ \hline

Tops       & Tops, Blouses, Dresses       & $556$      &  $926,033$  & $89,762$  \\ \hline
                            
All       & {}       & $1219$      & $3,480,575$  & $348,160$  \\
                            \hline
\end{tabular}
\label{tab:data_stats}
\end{table}

\noindent
\textbf{Top-Down Excitation Maps}.
Our goal is to interpret the deep model's predictions in order to explain the visual characteristics of fashion brands. Using the Excitation Backprop method~\cite{zhang:topdown}, we generate marginal probabilities on intermediate layers for the brand predicted with the maximum posterior probability, hence the name top-down. We assume the response in convolutional layers is positively correlated with their confidence of detection. This probabilistic framework produces well-normalized excitation maps efficiently via a single backward pass down to the target layer. We define two measures to encode the excitations:

\smallskip
\noindent{\textbf{Strength}}. Strength is calculated by computing the maximum over the excitation maps of a convolution layer. For every input image $x$,  we compute  excitation map $M_{k}(x)$ of every internal convolution unit $k$. 
We denote the excitation strength of convolutional layer by $S(x) = \max_{s\in{h_{k}}{{w_{k}}}}{E_{s}{(x)}}$, where $E_s(x) = \sum_{k=1\dots K}{M_{k}{(x)}}$
and $K$ is the total number of individual convolution units.

\smallskip
\noindent{\textbf{Extent}}. Extent is a measure to encode the spatial support of  high activations in excitation maps. Specifically, we first calculate the excitation map at every location $s$ across all units. Next we compute the ratio of locations where their excitation exceeds the mean value of all the excitations, represented by $T$. We define excitation extent of input image $x$ by $E(x) = \frac{1}{{h_{k}}{w_{k}}}\sum_{s\in{h_{k}}{{w_{k}}}}{\textbf{1}\left[E_s{(x)} > T\right]}$.

\smallskip
\noindent
\textbf{Discriminability}.
We aim to find units/neurons that often get high excitement values corresponding to a given brand. For every convolutional unit, we calculate the maximum value over the entire excitation map for every image $I$ and compute two distributions for positive $P^{+}$ and negative $P^{-}$ images associated to a brand $b$. For every  unit $k$, we compute the symmetric KL divergence~\cite{vittayakorn:automaticattribute}: $D_k(b|I) = KL(P^+ || P^-) + KL(P^- || P^+)$. The units that maximize the distance between the class conditional probabilities are deemed to have higher discriminability.

\section{Experiments}
\subsection{Brand Representations}
How do clothing brands make their products stand out among others? What do fashion designers do to appeal to shoppers? In order to answer these questions we begin by exploring  two ends of the spectrum of visual branding: brands which make themselves stand out through a localized mark, sign or logo, e.g. Chanel bags or Polo Ralph Lauren Shirts, and brands that convey their message via a spread design using colors and patterns, think colorful Vera Bradley or woven leather Bottega Veneta bags. In the following, we conduct our experiments on the bags category as it depicts a wide range of brand visualization strategies and receives the best classification score  among the categories in our dataset.

\begin{figure}
\centering
\includegraphics[width=1.0\textwidth]{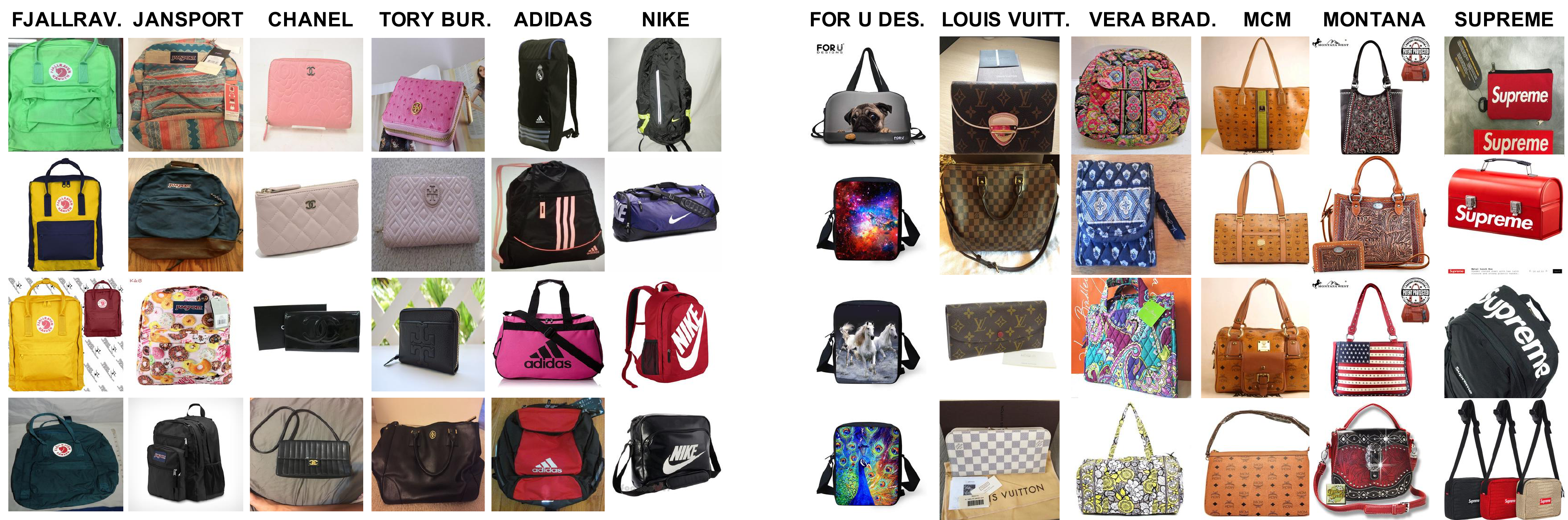}
\caption{Left: Brands with high Strength. Examples of bags from brands with high excitation strength are shown. All brands show concentrated logos or printed brand names. Right: Brands with high Extent. Examples of bags with high excitation extent are shown. Some brands print a large graphic on their products while other have a repeated pattern or logo. Composition of image can affect the extent signal as shown in the examples in the last column with large, repeated or close-up logos.}
\label{fig:experiments_top_brands_strength_extent}
\end{figure}

\smallskip
\noindent\textbf{Strength.}
Figure~\ref{fig:experiments_top_brands_strength_extent} depicts brands that obtain the highest excitation strengths. For each brand we compute median of the predicted strength across all samples in the test set. We find that brands such as Fjallraven, Jansport and Coach, design their bags with a unique logo or mark their goods with their brand name.

\smallskip
\noindent\textbf{Extent.}
What are the brands that are not as invested in logos and instead are interested to convey their message via unique patterns? Figure~\ref{fig:experiments_top_brands_strength_extent} depicts brands with the highest excitation extent values. For each brand, we find the extent decile to which it belongs to by computing the median of all extent values in the test set. Brands such as For U Designs, print large graphics of animals, nature or galaxy on their bags. Louis Vuitton makes their products remarkably recognizable via a unique checkered pattern or the famous repeated LV monogram. Vera Bradley is filled with colorful floral and paisley patterns and MCM repeats it's logo across a large region of the product. We also see how composition of images in a brand can contributes to large extent levels. The illustrated examples of Supreme brand are photographed in close-up and show the brand name in large size which leads to expanded excitations.

\begin{figure}[t]
\centering
\includegraphics[width=1.0\textwidth]{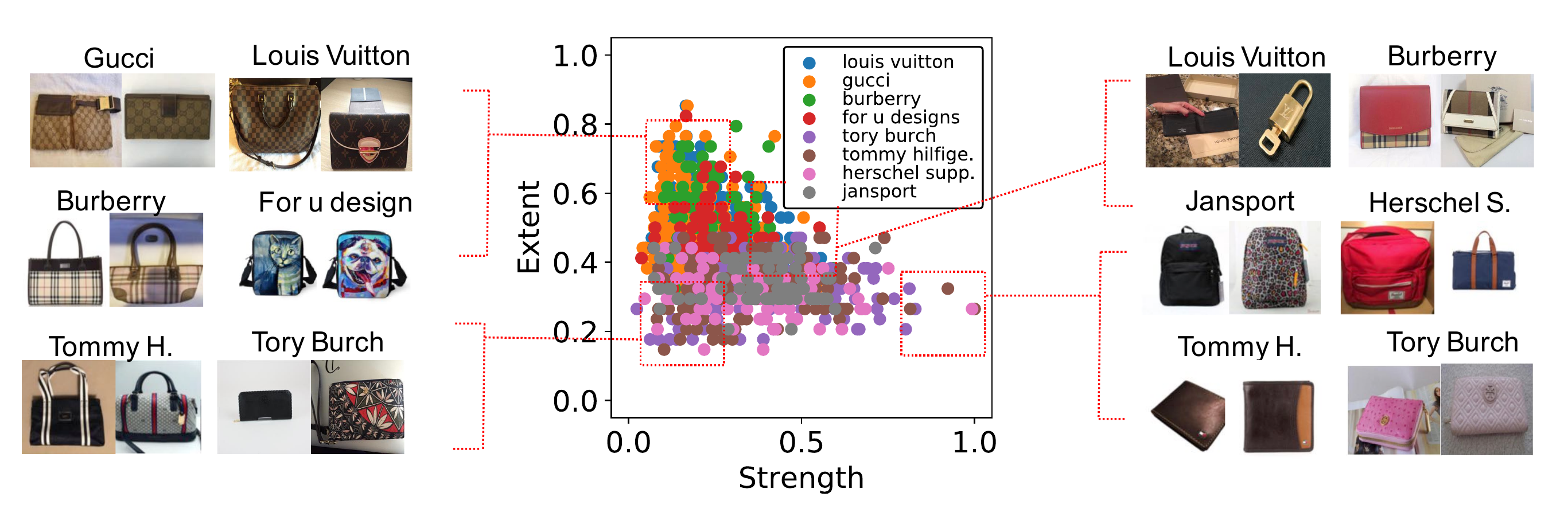}
\caption{Extent vs Strength. Depicts the transition of brands across the spectrum. We show samples that fall across the spectrum of extent and strength. Samples with high extent show a repeated texture or a large pattern. Items with high strength show a localized mark, logo or brand name. Hard examples to recognize such as Tommy Hilfiger bags that only show a specific type of stripes require specialized neurons to detect them.}
\label{fig:extent_vs_strength_bags}
\end{figure}

\smallskip
\noindent\textbf{Extent vs. Strength.}
Next we explore the space of Extent and Strength jointly. We ask, which brands have high extent but no single strong excitation value in their maps or vice versa? Are there examples that have both high or low extent and strengths? In order to answer these interesting questions, we plot the samples of top and bottom brands with samples falling in the highest and lowest extents in  Figure~\ref{fig:extent_vs_strength_bags}. We find that brands such as Burberry, Gucci or For U Designs are concentrated in the higher half of the spectrum, while logo-heavy brands such as Tommy Hilfiger, Tony Burch and Herschel Supply Co. bags are spread along the strength axis with low extent values. Interestingly, we observe that the model picks up signals, however weak, in the straps of Tommy Hilfiger totes, striped in iconic colors of Tommy Hilfiger. Comparing Tory Burch bags along the spectrum, the logos are hard to capture in the examples falling on the lower side of the strength axis while images of the same brand with high strength show fully visible logos. We also probe the middle region and observe an interesting phenomena. Louis Vuitton and Burberry images that fall in between, show a mix of logo monograms and brand names instead of just patterns.

\subsection{Versatility of Convolutional Units}
\begin{figure}
\centering
\caption{Left: Top activated neurons for Adidas and Burberry brands. Three top neurons specialized in recognizing in (a) Adidas and (b) Burberry. First column show a specialized neuron for recognizing the logo associated with a brand name, second column is specialized for three stripes, last column recognizes the smaller scale logo. First column in (b) recognizes vertical and horizontal stripes, second shows examples for neuron specialized in diagonal patterns and lastly is the logo detector for Louis Vuitton. Right: Specialist  vs generalist units.}
\includegraphics[width=1.0\textwidth]{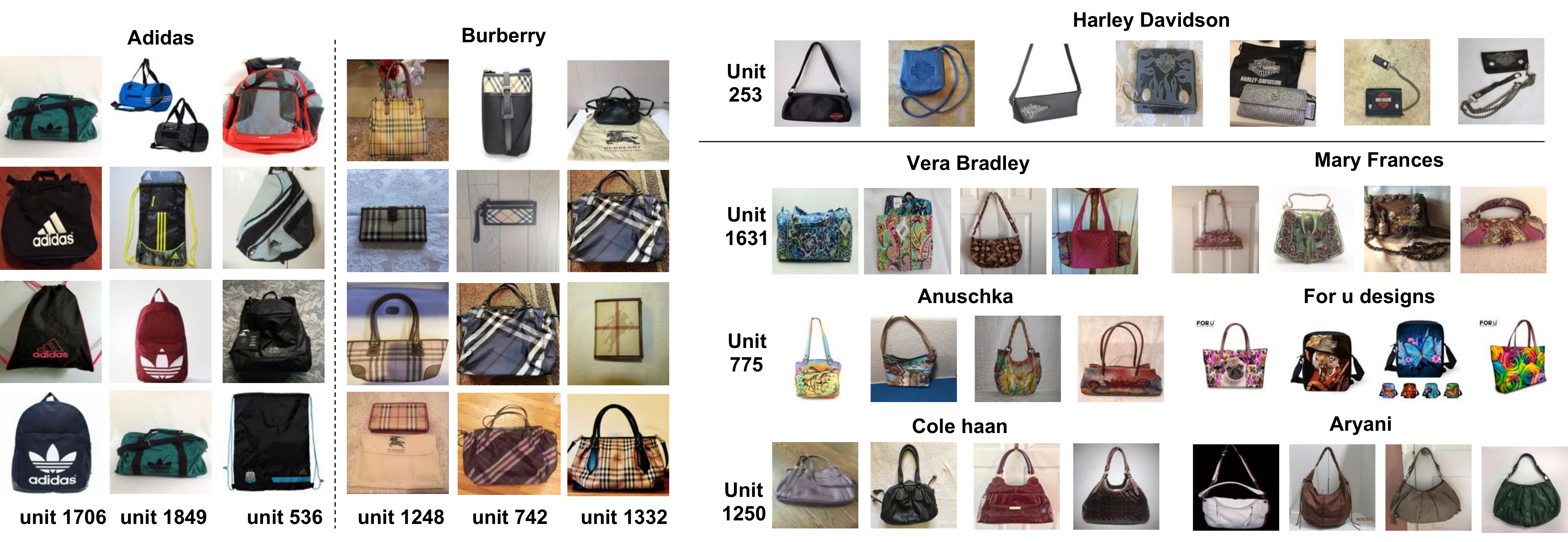}
\label{fig:experiments_specialist_generalist}
\end{figure}

Next, we go one step deeper and rank the convolutional units/neurons of the layer for each brand based on symmetric KL divergence score. We observe some neurons detect complex entangled concepts while others are more interpretable and specialized towards disentangled visual features. Figure~\ref{fig:experiments_specialist_generalist} left, shows top detection examples of such neurons for two sample brands. Some Adidas neurons detect the logos while others are specialized to detect vertical or horizontal stripes. For Burberry, we find units that detect diagonal or straight patterns, while another unit is more sensitive to the horse rider in the Burberry knight logo.

We further investigate ``specialist" vs. ``generalist" units. We compute the number of brands activated for each unit. Specialists units are activated for only one or few brands. Figure~\ref{fig:experiments_specialist_generalist} right, shows examples of specialist units and the brands they activate. Unit $253$ is an expert only in detecting the Harley Davidson logo, which is unique and can happen in many locations over the object and requires its own specialized unit. Meanwhile, units $1631$ and $770$ detect floral and natural patterns that are more general and shared among brands such as Vera Bradley and Mary Frances. Unit $1250$ is specialized to detect hobo-shaped bags with a large crescent-shaped bottom and a shoulder strap that represents multiple brands. By analyzing the space of specialist units we can discover unique visual expressions that sets a brand apart. Generalist units point us to features shared by several brands.

\begin{figure}
\centering
\includegraphics[width=1.0\textwidth]{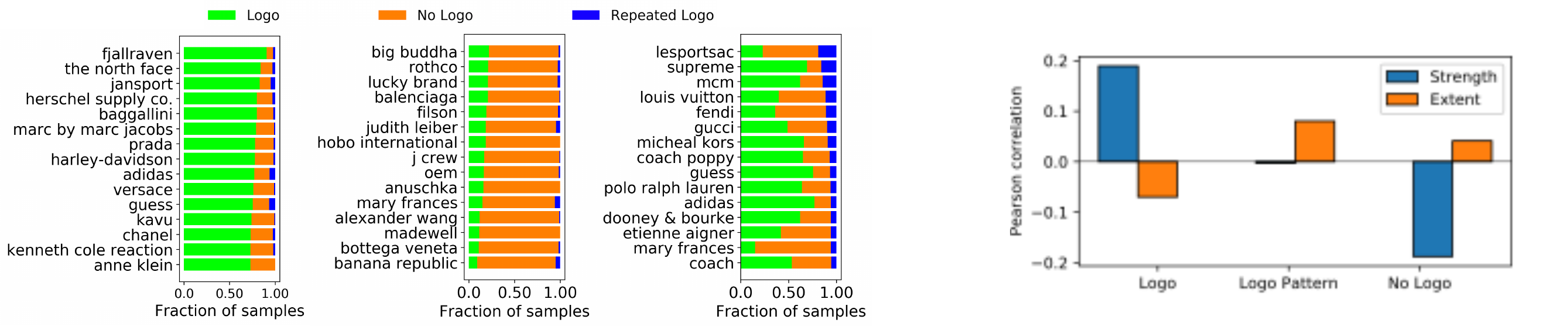}
\caption{Left: Individual brands ranked in the logo visibility spectrum.  Brands are sorted based on their fraction of samples labeled by humans as (i) Logo, (ii) No Logo or (iii) Repeated Logo. Right: Pearson correlation of Strength and Extent of excitations with brand visibility variations.}
\label{fig:human_exp}
\end{figure}

\subsection{Human Experiment}
We conduct a human study asking 5 subjects on Amazon Mechanical Turk to label each product image in the bags category according to the visibility of the logo into one of three groups: (i) Logo (ii) Repeated Logo, when a pattern of repeated logos or monogram and (iii) No Logo. $46\%$ contain a visible logo and $51\%$ contain no logo. This is particularly interesting as a recent study shows that one third of the handbags purchased in the U.S. did not have a visible logo~\cite{cnbc:nologo}. The classifier correctly predicts $65.01\%$, $68.67\%$ and $54.46\%$ of the brands in groups (i), (ii) and (iii) respectively. This is significant, given that group (iii) constitutes the majority of the dataset and confirms that deep classifiers learn unique visual characteristics of all three groups.

\smallskip
\noindent
\textbf{Logo Visibility in Brands.}
We further study individual brands by ranking them based on the ratio of samples that fall into each of the three logo visibility groups. Figure~\ref{fig:human_exp}, shows that brands such as Fjallraven and The North Face are logo-based. On the other hand, Lucky Brand does not depend on logo. Instead, they claim to give your look ``the added flare" by embellishments such as fringe and embroidered detailing. Fendi and MCM opt in to repeat their logo in their design. In fact, the ``Shopper" totes from MCM are reversible with their logo printed on both inside and outside! 

\smallskip
\noindent
\textbf{Correlation with Strength and Extent}
Finally we compute the Pearson correlation between the predicted strength and extent of the excitation maps and report the results in Figure~\ref{fig:human_exp}. We find that excitation strength has  strong correlation with samples depicting a logo while extent has a negative correlation with logo-oriented products. Products with repeated logo produce scattered signals with low strength and high extent. For an item with no logo, the network needs to aggregate signals from various spatial locations and hence it is positively correlated with extent and negatively correlated with strength.

\noindent
\textbf{Conclusion}.
In this work, we quantify the deep representations to analyze and interpret visual characteristics of fashion brands. We find units that are specialized to detect specific brands as well as versatile units that detect shared concepts. A human experiment confirms the proposed measures are aligned with human perception.

\bibliographystyle{splncs04}
\bibliography{mybibliography}

\begin{thebibliography}{10}
\providecommand{\url}[1]{\texttt{#1}}
\providecommand{\urlprefix}{URL }
\providecommand{\doi}[1]{https://doi.org/#1}

\bibitem{agrawal:analyzingtheperformance}
Agrawal, P., Girshick, R., Malik, J.: Analyzing the performance of multilayer
  neural networks for object recognition. European Conference on Computer
  Vision  (2014)

\bibitem{alhalah:fashionforward}
Al-Halah, Z., Stiefelhagen, R., Grauman, K.: Fashion forward: Forecasting
  visual style in fashion. International Conference on Computer Vision  (2017)

\bibitem{bau:networkdissection}
Bau, D., Zhou, B., Khosla, A., Oliva, A., Torralba, A.: Network dissection:
  Quantifying interpretability of deep visual representations. Conference on
  Computer Vision and Pattern Recognition  (2017)

\bibitem{chen:describing}
Chen, H., Gallagher, A., Girod, B.: Describing clothing by semantic attributes.
  European Conference on Computer Vision  (2012)

\bibitem{chen:deepdomain}
Chen, Q., Huang, J., Feris, R., Brown, L.M., Dong, J., Yan, S.: Deep domain
  adaptation for describing people based on fine-grained clothing attributes.
  Conference on Computer Vision and Pattern Recognition  (2015)

\bibitem{escorcia:ontherelationship}
Escorcia, V., Niebles, J.C., Ghanem, B.: On the relationship between visual
  attributes and convolutional networks. Conference on Computer Vision and
  Pattern Recognition  (2015)

\bibitem{gong:lookinto}
Gong, K., Liang, X., Zhang, D., Shen, X., Lin, L.: Look into person:
  Self-supervised structure-sensitive learning and a new benchmark for human
  parsing. Conference on Computer Vision and Pattern Recognition  (2017)

\bibitem{he:deepresidual}
He, K., Zhang, X., Ren, S., Sun, J.: Deep residual learning for image
  recognition. Conference on Computer Vision and Pattern Recognition  (2016)

\bibitem{kiapour:wtb}
Kiapour, M.H., Han, X., Lazebnik, S., Berg, A.C., Berg, T.L.: Where to buy it:
  Matching street clothing photos in online shops. International Conference on
  Computer Vision  (2015)

\bibitem{kiapour:hipsterwars}
Kiapour, M.H., Yamaguchi, K., Berg, A.C., Berg, T.L.: Hipster wars: Discovering
  elements of fashion styles. European Conference on Computer Vision  (2014)

\bibitem{nologo}
Klein, N.: No Logo: Taking Aim at the Brand Bullies. Random House of Canada,
  Picador (1999)

\bibitem{krizhevsky:imagenet}
Krizhevsky, A., Sutskever, I., Hinton, G.E.: Imagenet classification with deep
  convolutional neural networks. Information Processing Systems  (2012)

\bibitem{liu:deepfashion}
Liu, Z., Luo, P., Qui, S., Wang, X., Tang, X.: Deepfashion: Powering robust
  clothes recognition and retrieval with rich annotations. Conference on
  Computer Vision and Pattern Recognition  (2016)

\bibitem{liu:fashionlandmark}
Liu, Z., Yan, S., Luo, P., Wang, X., Teng, X.: Fashion landmark detection in
  the wild. European Conference on Computer Vision  (2016)

\bibitem{cnbc:nologo}
Money, C.M.: No logo: Why un-branded luxury goods are on the rise  (2016),
  \url{https://www.cnbc.com/2016/11/28/no-logo-why-un-branded-luxury-goods-are-on-the-rise.html}

\bibitem{ozeki:understandingconvolutional}
Ozeki, M., Okatani, T.: Understanding convolutional neural networks in terms of
  category-level attributes. Asian Conference on Computer Vision  (2014)

\bibitem{ruth:interpretable}
Ruth, F., Vedaldi, A.: Interpretable explanations of black boxes by meaningful
  perturbation. International Conference on Computer Vision  (2017)

\bibitem{selvaraju:gradcam}
Selvaraju, R.R., Cogswell, M., Das, A., Vedantam, R., Parikh, D., Batra, D.:
  Grad-cam: Visual explanations from deep networks via gradient-based
  localization. International Conference on Computer Vision  (2017)

\bibitem{springenberg:striving}
Springenberg, J.T., Dosovitskiy, A., Borx, T., Riedmiller, M.: Striving for
  simplicity: The all convolutional net. arXiv preprint arXiv:1412.6806  (2014)

\bibitem{veit:learningvisual}
Veit, A., Kovacs, B., Bell, S., McAuley, J., Bala, K., Belongie, S.: Learning
  visual clothing style with heterogeneous dyadic co-occurrences. International
  Conference on Computer Vision  (2015)

\bibitem{vittayakorn:automaticattribute}
Vittayakorn, S., Umeda, T., Murasaki, K., Sudo, K., Okatani, T., Yamaguchi, K.:
  Automatic attribute discovery with neural activations. European Conference on
  Computer Vision  (2016)

\bibitem{yamaguchi:paperdoll}
Yamaguchi, K., Kiapour, M.H., Ortiz, L.E., Berg, T.L.: Retrieving similar
  styles to parse clothing. IEEE Transactions on Pattern Analysis and Machine
  Intelligence  (2014)

\bibitem{yamaguchi:mix}
Yamaguchi, K., Okatani, T., Sudo, K., Murasaki, K., Taniguchi, Y.: Mix and
  match: Joint model for clothing and attribute recognition. BMVC  (2015)

\bibitem{transferlearning}
Yosinski, J., Clune, J., Bengio, Y., Lipson, H.: How transferable are features
  in deep neural networks? Advances in Neural Information Processing Systems
  (2014)

\bibitem{zeiler:visualizing}
Zeiler, M.D., Fergus, R.: Visualizing and understanding convolutional networks.
  European Conference on Computer Vision  (2014)

\bibitem{zhang:topdown}
Zhang, J., Lin, Z., Brandt, J., Shen, X., Sclaroff, S.: Top-down neural
  attention by excitation backprop. European Conference on Computer Vision
  (2016)

\bibitem{zhou:objectdetectors}
Zhou, B., Khosla, A., Lapedriza, A., Oliva, A., Torralba, A.: Object detectors
  emerge in deep scene cnns. International Conference on Learning
  Representations  (2015)

\end{thebibliography}

\end{document}